%% file: main.tex
\input{packages}

\title{Magpie: Real-Time World Renderer for Interactive Games}

\input{authors}

\begin{document}

\setcounter{secnumdepth}{2}

\makeatletter
\setlength\titlebox{6.5in}
\def\@maketitle{
  \newcounter{eqfn}\setcounter{eqfn}{0}
  \vbox to \titlebox {
    \hsize\textwidth
    \linewidth\hsize
    \vskip 0.625in minus 0.125in
    \centering
    \makebox[\textwidth][c]{%
      \raisebox{-0.18\height}{\includegraphics[height=0.48in,keepaspectratio]{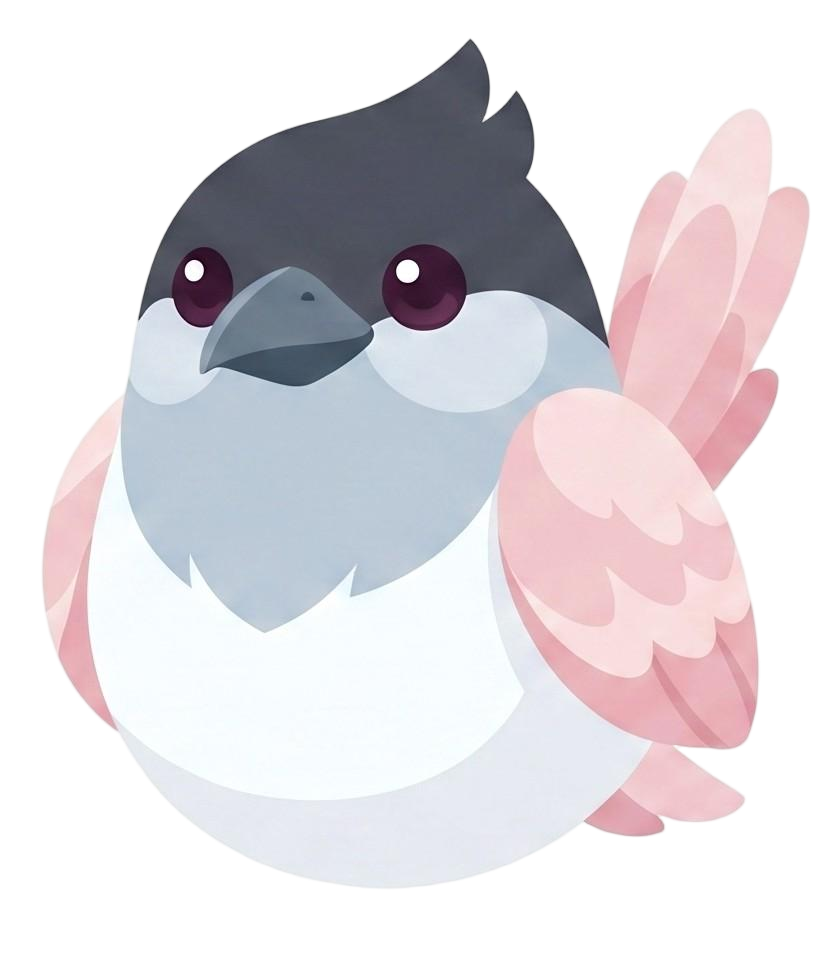}}%
      \hspace{0.12in}%
      {\Large\bfseries \@title}%
    }\par
    \vskip 0.1in plus 0.5fil minus 0.05in
    {\Large{\textbf{\@author\ifhmode\\\fi}}}
    \vskip .2em plus 0.25fil
    {\normalsize \affiliations_\ifhmode\\\fi}
    \vskip 1em plus 2fil

    \begin{center}
        \centering
        \includegraphics[width=0.95\textwidth]{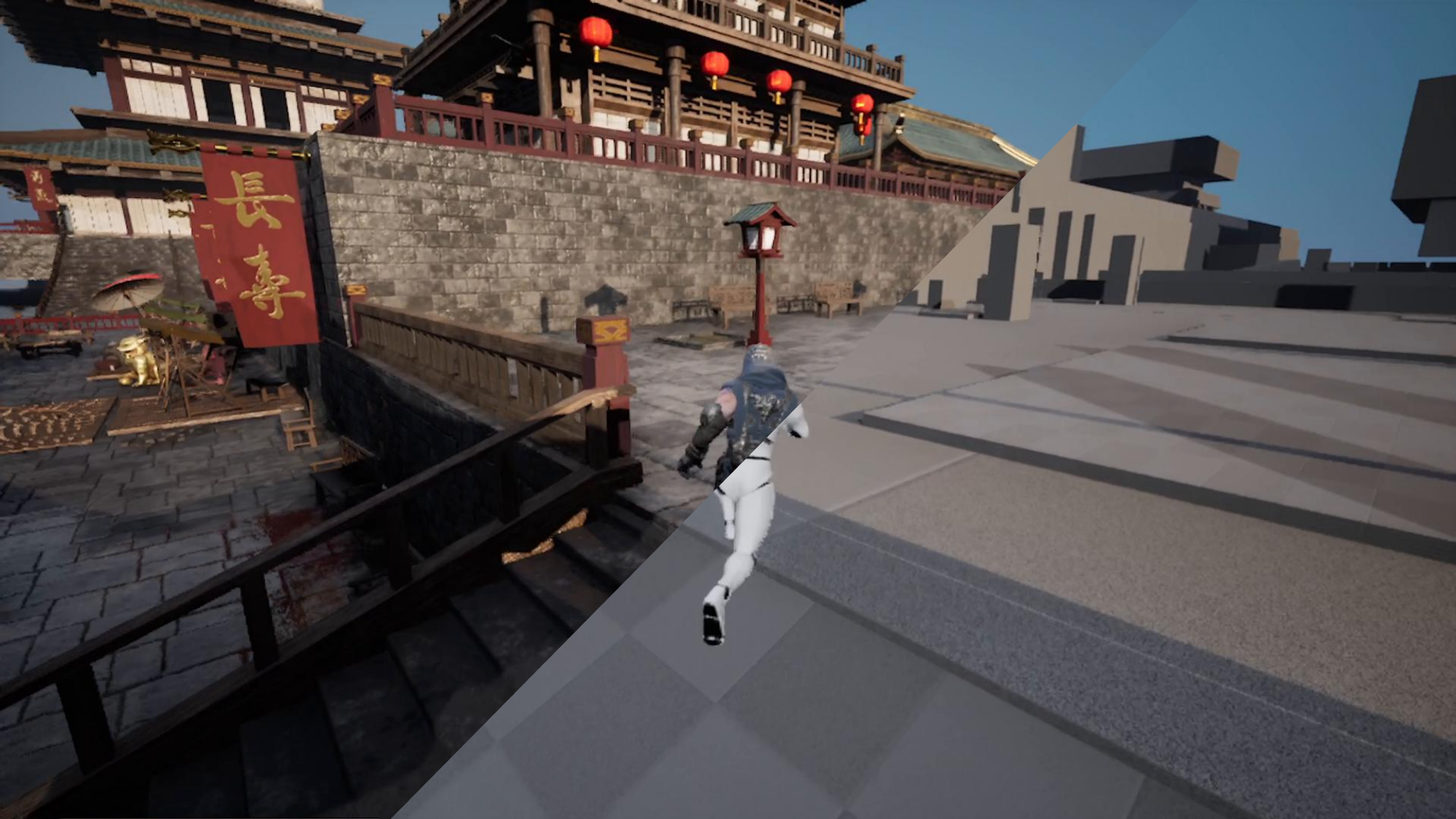}
        \captionof{figure}{\textbf{From white-box gameplay to generative rendering.} Magpie hopes to keep layout, rules, and runtime state inside the Game Engine while the Render Server transforms the aligned white-box observation into a visually complete frame.}
        \label{fig:teaser}
    \end{center}
  }
}
\makeatother

\maketitle

\let\thefootnote\relax\footnotetext{\textsuperscript{*}Equal contribution. \textsuperscript{\dag}Corresponding author.\\ \hspace*{1.8em}\href{https://zhanxy.xyz/Magpie-website}{https://zhanxy.xyz/Magpie-website}}

\input{Sections/0_abstract}
\input{Sections/1_introduction}
\input{Sections/2_related_work}

\input{Sections/3_overview}
\input{Sections/4_data_engine}
\input{Sections/5_pipeline}
\input{Sections/6_training}
\input{Sections/7_performance}

\input{Sections/8_results}

\input{Sections/9_conclusion}

\bibliography{arxiv}

\end{document}

%% file: packages.tex
\documentclass[letterpaper]{article} 
\usepackage{arxiv}  

\usepackage{times}  
\usepackage{helvet}  
\usepackage{courier}  
\usepackage[hyphens]{url}  
\usepackage{graphicx} 
\usepackage{natbib}  
\usepackage{caption} 
\usepackage{multirow}
\usepackage{booktabs}      
\usepackage{subcaption}

\usepackage{algorithm}
\usepackage{xcolor}
\usepackage{colortbl}
\usepackage{amsmath, amssymb, amsthm}
\usepackage{cuted}

\definecolor{refpurple}{RGB}{90,60,140}

\definecolor{tab1}{RGB}{255,153,153}
\definecolor{tab2}{RGB}{255,204,153}
\definecolor{tab3}{RGB}{255,246,178}

\usepackage{newfloat}
\usepackage{listings}
\DeclareCaptionStyle{ruled}{labelfont=normalfont,labelsep=colon,strut=off} 
\floatstyle{ruled}
\newfloat{listing}{tb}{lst}{}
\floatname{listing}{Listing}
\usepackage{bibentry}

\usepackage{hyperref}
\hypersetup{
  colorlinks=true,
  linkcolor=refpurple,
  citecolor=refpurple,
  urlcolor=refpurple,
  unicode=true
}

%% file: authors.tex
\author{
    Xiaoyu Zhan$^{12*}$,
    Xinyu Wang$^{1*}$,
    Xiaohong Zhang$^{12*}$,
    Huanjie Zhu$^{1*}$, \\
    Tengjiao Sun$^{13}$,
    Pengcheng Fang$^{13}$,
    Jiaxing Yu$^{12}$,
    Yanwen Guo$^{2}$, 
    Dongjie Fu$^{1\dag}$
}
\affiliations {
    {\Large 
        $^{1~}$Mogo AI Ltd. \quad 
        $^{2~}$Nanjing University \quad
        $^{3~}$University of Southampton
    }
}

%% file: Sections/0_abstract.tex
\begin{abstract}
Modern game development relies heavily on conventional graphics pipelines. High-quality visual content requires modeling, material authoring, animation, lighting, effects, and runtime optimization, making asset production expensive and extending the development cycle of game prototypes. Recently, video foundation models are beginning to change film and video production, but games differ from linear media, they require not only continuous and realistic imagery, but also stable and reproducible gameplay rules, object states, and interaction outcomes. We present \textbf{Magpie}, a real-time generative world-rendering system for interactive games. Magpie separates gameplay execution from visual generation. Designers define scenes and rules in a game engine. At runtime, the Game Engine resolves player actions and maintains world state, while an independent Render Server generates visual output from white-box frames produced by the engine. The Render Server receives a text prompt and a first-frame image specifying the visual style only during initialization. During subsequent interaction, white-box frames serve as the continuing denoising condition, and camera poses retrieve historical frames relevant to the current viewpoint. Player actions, state variables, object properties, and event signals remain in the Game Engine and will not be passed directly to the Render Server. The generative model is therefore responsible for visual presentation, while the Game Engine continues to execute gameplay rules and control state. To train Magpie, we manually collect approximately 300 hours of interactive video in Unreal Engine scenes. The data covers basic locomotion, viewpoint changes, driving, sitting, collision interactions, and idle states, with synchronized high-fidelity renderings, white-box renderings, camera poses, and structured interaction records. Magpie provides a system-level implementation path for applying generative models to real-time game rendering. It preserves gameplay designability and reproducibility, and reduces the dependence of early game prototypes on complete visual assets.
\end{abstract}

%% file: Sections/1_introduction.tex
\section{Introduction}
\label{sec:intro}

\leavevmode\hspace*{\parindent}%
Conventional graphics pipelines are constrained by both asset-production cost and runtime performance. High-quality game renderings depends on a complete workflow spanning modeling, texturing, materials, rigging, animation, lighting, and performance optimization, often requiring sustained collaboration among many specialists. Even with ray tracing or high-quality rasterization on consumer GPUs, complex materials, illumination, and environmental phenomena remain approximations. Compared with the quality of visual-asset production, gameplay and experience design more directly constitute the core value of a game. While these production costs and technical barriers make it difficult for many creators to produce a visually polished game independently. As technology advances, the recent visual capabilities of video foundation models offer a new means of reducing the cost of game rendering production. These models are already affecting film production, which likewise depends on asset creation and rendering workflows. When will games follow suit?

Games, however, are not videos that simply unfold over time. They are designed interactive systems in which players participate. Designers construct intended experiences through rules, numerical parameters, and state relationships. It does not require every session to produce the same outcome, but it requires the same actions and states to retain stable meanings under the specified rules. Only under this condition can mechanics be tested and adjusted, design knowledge be communicated accurately, and a local prototype be developed into a complete product.

Game engines provide an executable system description for such interactive experiences. Spatial constraints, object states, interaction rules, event conditions, and state transitions are maintained explicitly by the engine. Collision volumes, traversal permissions, cooldowns, hidden triggers, and progression flags leave little reliable evidence in rendered images and are therefore difficult to infer from video, but they can be represented directly in an engine. Designers can also use the engine to inspect state, modify rules, replay scenes, and compare versions, making it possible to analyze how a specific design change affects the resulting experience. The game engine is therefore not only the software environment in which a game runs, but also the foundation of gameplay design and iteration. During the initial phase of gameplay design, designers can forego high-fidelity assets and instead use coarse models, which are adequate for constructing a functional prototype.

In order to ensure the visual experience of players, conventional real-time graphics pipelines must approximate real-world material appearance and light transport within a limited frame time. Physically based shading, baked lighting, reflection approximations, and post-processing can efficiently represent many common effects, but complex materials, indirect illumination, translucent media, changing weather, and surface interactions remain difficult to reproduce fully. High-quality imagery also depends on a complete asset-production process, including modeling, topology, UVs, textures, material calibration, rigging, animation, lighting, effects, and multiple levels of optimization. The resulting imagery remains constrained by asset coverage and runtime budgets. A game prototype may therefore have complete gameplay but still fall short of conveying the designer’s intended experience, as finishing the visual assets will demand extra time and money.

Video generation models use a different approach to visual modeling. Large diffusion Transformers such as Wan and Seedance learn correspondences among appearance, motion, and multimodal conditions~\citep{wan2025,seedance2026,agarwal2026cosmos,hacohen2026ltx,chen2026skyreels}. These models can represent highly realistic materials, lighting, environmental detail, and object motion, while adapting to different visual styles. Once trained, the same model can produce multiple visual styles from simple conditions, making the marginal cost of exploring a new appearance much lower than rebuilding a production asset set. Their present strength, however, lies primarily in generating observable imagery rather than maintaining game rules and state. A video model alone cannot reliably guarantee consistent interaction rules or provide designers with precise state control and debugging tools. Apart from this, applying them to real-time games additionally requires timely responses to player input, preservation of necessary history during continuous interaction, and stable adherence to scene structure. 

Motivated by this division of responsibilities, we present \textbf{Magpie}, a real-time generative world-rendering system for controllable interactive games. Gameplay Design defines scenes and rules; the Game Engine resolves player actions, executes those rules, and maintains world state; and the Render Server hosts the generative world renderer and produces visual output. A text prompt and first-frame image initialize the Render Server. Once interaction begins, white-box frames generated by the engine are the only continuing denoising condition, while camera poses retrieve historical frames related to the current viewpoint. Player actions, game state, object properties, and event signals remain in the Game Engine and are never passed directly to the Render Server.

Within this system, the Game Engine determines what happens in the game world and stores gameplay-related state. Designers can inspect and modify that state and reproduce the same interaction under controlled conditions. The Render Server generates the corresponding imagery from structural conditions supplied by the engine, but does not participate in judging or executing gameplay rules. This separation preserves the designability and reproducibility of the original game system while allowing a generative model to participate in real-time visual construction.

For players, Magpie can generate different visual styles, materials, lighting, environmental details, and secondary motion for the same playable scene, enabling more flexible visual expression and personalized experiences. For developers, Magpie can reduce the dependence of the prototype stage on complete visual assets. A white-box scene can approach the intended visual effect before detailed modeling, texturing, rigging, lighting, and performance optimization are complete. Developers can therefore evaluate gameplay and experience design earlier while continuing to control rules, state, and iteration directly through the Game Engine. Within this framework, generated imagery serves visual presentation, while its foundation remains the game system explicitly constructed by designers.

%% file: Sections/2_related_work.tex
\section{Related Work}
\label{sec:related_work}

\begin{figure*}[t]
  \centering
  \includegraphics[width=0.97\textwidth]{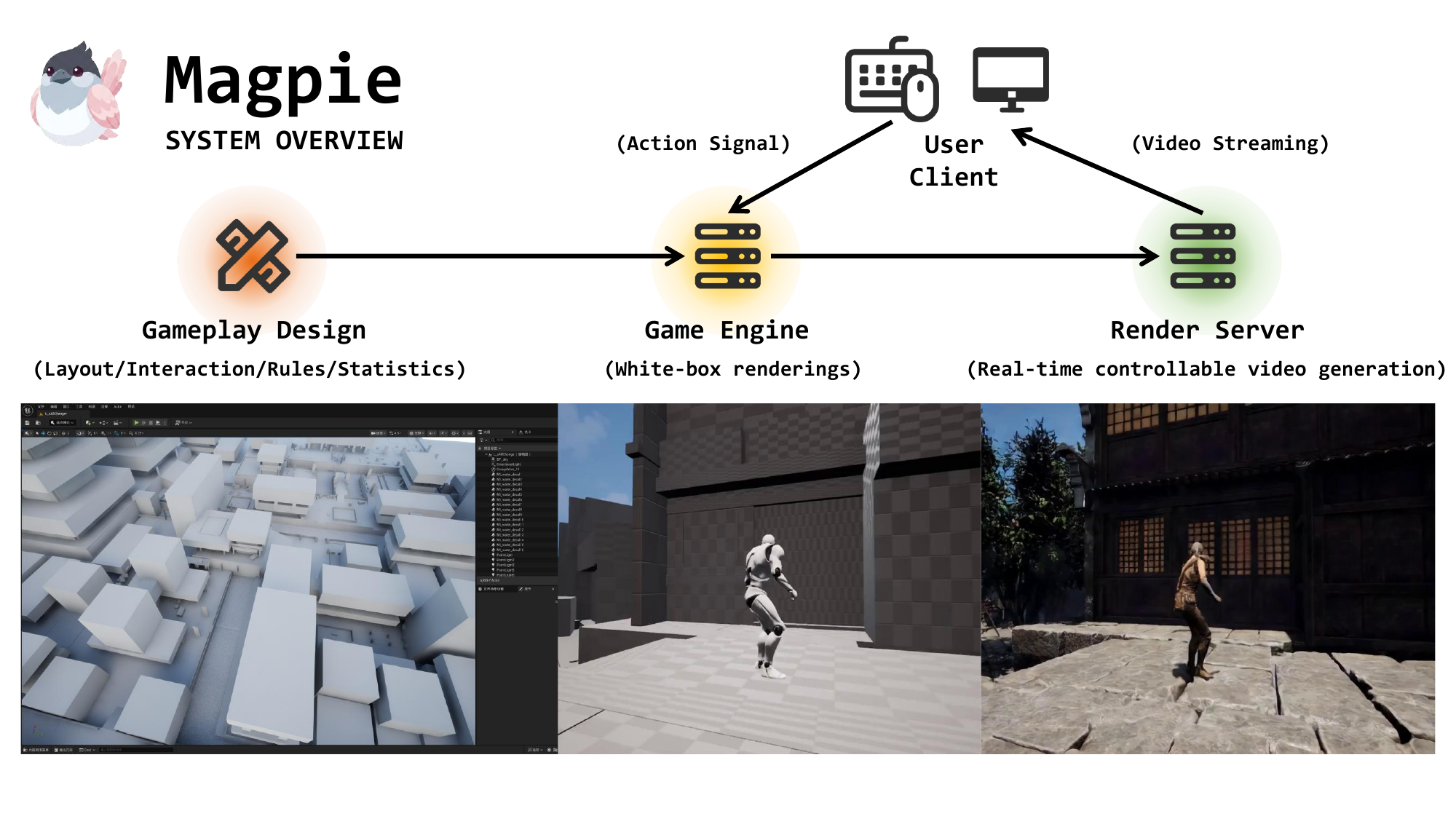}
  \caption{\textbf{Magpie system overview.} Designers define layouts, interactions, rules, and numerical parameters in the game engine. The User Client forwards player actions to the Game Engine, which executes gameplay and emits white-box observations. The independent Render Server converts those observations into generated video and streams the result back to the client. Gameplay state and rule execution remain inside the engine.}
  \label{fig:system_overview}
\end{figure*}

\subsection{Real-Time Video Generation}

\leavevmode\hspace*{\parindent}%
Real-time video generation demands causal inference with bounded sampling and context overheads that do not escalate over time. Existing studies approach this primarily via sampling acceleration, low-precision deployment, and long-horizon memory.

Sampling acceleration has progressed from general one-step generation to causal distillation for autoregressive video. Distribution matching~\citep{yin2024dmd,yin2024improved}, adversarial distillation~\citep{sauer2024adversarial}, and score-regularized consistency~\citep{zheng2026rcm} compress multi-step diffusion into short sampling trajectories, while MeanFlow~\citep{geng2026mean} directly learns a one-step average velocity field. AnyFlow~\citep{anyflow2026} learns flow-map transitions over arbitrary time intervals within a single distillation procedure, enabling one video model to support both few-step and multi-step sampling and to trade inference cost for generation quality at deployment. For continuous rollout, Self-Forcing~\citep{selfforcing2025} reduces the discrepancy between teacher-forced training and inference on generated history. Causal Forcing~\citep{causalforcing2026} and Causal Forcing++~\citep{causalforcingplusplus2026} further align causal teachers and students for few-step chunk-wise and frame-wise generation. LongLive~\citep{longlive2025} and Helios~\citep{helios2026} integrate these objectives with causal architectures and compressed context to support minute-scale generation at interactive throughput.

Low-precision deployment reduces the remaining arithmetic and memory costs. Quantization-aware distillation~\citep{xin2026quantization} recovers accuracy in quantized students, while LongLive-2.0~\citep{chen2026longlive2} combines NVFP4 execution, quantized key and value caches, parallel inference, and asynchronous decoding. These results complement the INT8 deployment of LongLive~\citep{longlive2025} and show that model compression and runtime scheduling should be optimized jointly.

Long-horizon generation additionally requires bounded visual memory. Context-as-Memory~\citep{contextasmemory2025} retrieves frames through camera field-of-view overlap, and WorldKV~\citep{worldkv2026} retrieves and compresses evicted attention states. Echo-Infinity~\citep{echoinfinity2026} instead consolidates history into learned fixed-cost queries, while Mirage~\citep{latentspatialmemory2026} stores diffusion features in a persistent three-dimensional cache. These methods retain evidence for revisitation without allowing context cost to grow with rollout length.

Together, distillation, quantization, and bounded memory provide the computational basis for persistent video rendering. Magpie adopts these techniques for visual synthesis while retaining rule execution and authoritative world state in the Game Engine.

\subsection{Real-Time Interactive Video Generation}

\leavevmode\hspace*{\parindent}%
Real-time interactive video requires the network to perform streaming generation based on received control signals. Existing frameworks mainly adopt two types of interaction logic. One is real‑time control from keyboard inputs and camera poses. The other supports real‑time intervention from text prompts.

Keyboard-conditioned and camera-conditioned frameworks represent control as discrete actions, viewpoint motion, or geometric pose. Matrix-Game~\citep{matrixgame2025} and Matrix-Game 2.0~\citep{he2025matrix2} inject frame-level keyboard and mouse inputs, while WorldCam~\citep{worldcam2026} uses camera geometry for both immediate control and historical indexing. ActWorld~\citep{xiong2026actworld} and ABot-World-0~\citep{jiang2026abot} extend keyboard control from navigation to object and third-person character interaction.

Text-controlled systems allow semantic instructions to modify events or scene content during an ongoing rollout. DreamX-World~\citep{dreamxworld2026} combines camera navigation with promptable events, while LingBot-World 2.0~\citep{lingbotworld2_2026} supports diverse character actions and text-driven event generation over an unbounded interaction horizon. Compared with keyboard and pose conditions, text provides a broader control vocabulary but specifies timing and spatial consequences less directly.

Memory mechanisms preserve scene identity and geometry beyond the recent context window—features that are also crucial for real-time interactive generation. WorldPlay~\citep{worldplay2025} reconstructs relevant historical context, RELIC~\citep{hong2025relic} and Matrix-Game 3.0~\citep{matrixgame3_2026} compress camera-aware history, and Wonder~\citep{wonder2026} retrieves sparse geometric evidence for revisitation. Computational optimization addresses the latency and hardware cost of maintaining such rollouts. LingBot-World~\citep{team2026lingbot1} extends interaction duration at real-time rates, SANA-WM~\citep{zhu2026sana} combines hybrid linear attention with low-precision deployment, and MoWorld~\citep{moxin2026moworld} jointly optimizes data, distillation, and mixed-precision inference. ABot-World-0~\citep{jiang2026abot} further combines low-bit inference, a lightweight decoder, and memory-aware scheduling for desktop deployment.

There are also some studies focus on multi-agent generation. Solaris~\citep{savva2026solaris} generates synchronized Minecraft views, MIRA~\citep{hu2026mira} models physically coupled multiplayer interaction, and WanToFight~\citep{hu2026wantofight} studies real-time two-player control in a two-dimensional fighting game.

These frameworks improve interactive visual generation, but neither keyboard input nor text instruction guarantees compliance with authored gameplay rules. Magpie instead lets the Game Engine resolve actions, collisions, events, and persistent state before producing the white-box observation. 

\subsection{Controllable Video Generation}

\leavevmode\hspace*{\parindent}%
Controllable video generation connects semantic, geometric, photometric, and motion representations through a shared video prior. Related studies include unified multimodal generation, three-dimensional control, and generative rendering from graphics buffers or coarse simulations.

Unified models formulate several tasks within one generative space. UniVidX~\citep{chen2026unividx} supports conditional generation among RGB, intrinsic maps, and compositing layers, while GenCeption~\citep{wang2026video} transfers video generation priors to depth, normal, pose, segmentation, and keypoint prediction. Diffusion as Shader~\citep{diffusionasshader2025} unifies mesh rendering, camera control, motion transfer, and object manipulation through three-dimensional tracking videos. These methods replace fixed task-specific mappings with aligned multimodal representations.

Apart from these, geometry-aware control makes scene composition externally editable. CineMaster~\citep{cinemaster2025} uses object layouts, depth, labels, and camera trajectories, while Diffusion as Shader~\citep{diffusionasshader2025} uses tracking signals shared across control tasks. Generative rendering methods instead expose physical intermediate representations. DiffusionRenderer~\citep{liang2025diffusionrenderer} learns mappings between RGB video and G-buffers, LightCrafter~\citep{guo2026lightcrafter} refines physically based relighting proxies, and Generative World Renderer~\citep{generativeworldrenderer2026} uses synchronized game footage and multiple G-buffer channels.

Coarse-to-Real~\citep{coarsetoreal2026} translates coarse dynamic simulations into realistic videos while preserving layout, camera motion, and human trajectories. Magpie extends this principle to a closed interactive loop. The Game Engine resolves gameplay and emits white-box video, after which the generative renderer supplies appearance and fine-scale dynamics. The condition therefore carries an executed visual consequence, preserving a clear boundary between gameplay computation and visual synthesis.

%% file: Sections/3_overview.tex
\section{System Overview}
\label{sec:overview}

\leavevmode\hspace*{\parindent}%
Magpie is organized around a direct separation between gameplay execution and visual generation, as summarized in Figure~\ref{fig:system_overview}. Designers construct the scene and its rules in the game engine, while the User Client forwards player input to the Game Engine. The engine executes the rules, maintains world state, and produces a white-box observation after resolving each action.

The independent Render Server converts the white-box observation into generated video and streams the result back to the User Client. A text prompt and a first-frame image specifying the visual style establish appearance at initialization. During interaction, white-box frames are the only continuing denoising condition, while camera poses are used only to retrieve relevant visual history. Player actions, object properties, event signals, and hidden gameplay state remain inside the Game Engine.

This separation gives each component a clear responsibility: the Game Engine determines what happens, and the Render Server determines how the resolved outcome is presented. A visual generation error may reduce the quality or clarity of player feedback, but it cannot change collision logic, progression, or subsequent gameplay decisions. By using a video generation model to transform white-box observations directly into final rendered imagery, Magpie bypasses much of the complex asset-production process that would otherwise be required to turn a functional prototype into a visually representative experience. 

%% file: Sections/4_data_engine.tex
\section{Magpie Data Engine}
\label{sec:data_engine}

\begin{figure*}[t]
  \centering
  \includegraphics[width=0.97\textwidth]{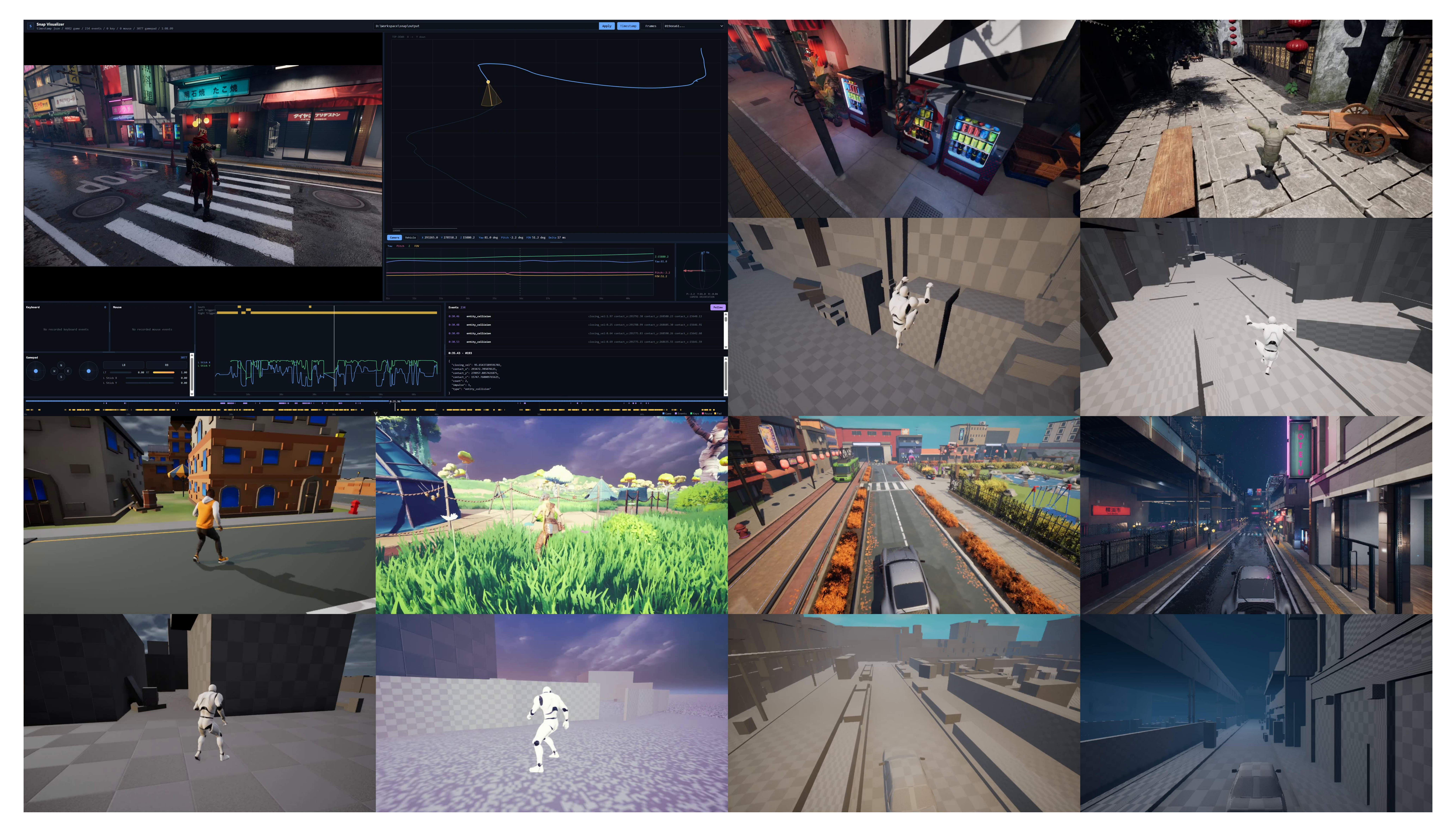}
  \caption{\textbf{Human-operated data collection and paired training examples.} The capture interface (top left) synchronizes operator inputs, camera trajectories, engine events, and timestamps. The remaining examples illustrate the diversity of manually played scenes and the spatially aligned high-fidelity/white-box pairs.}
  \label{fig:data_engine}
\end{figure*}

\leavevmode\hspace*{\parindent}%
The data engine is designed to teach visual rendering rather than gameplay execution. Rules and state transitions are authored and executed in the game engine, while training data provides time-aligned supervision for converting engine output into generated video. Each capture contains a high-fidelity target stream, a synchronized white-box stream, camera poses for historical retrieval, and text describing the appearance established at initialization. The capture format records more information than the current model consumes. Operator inputs, collisions, state transitions, and event records are preserved as synchronized metadata, but they are not used as training conditions or targets in this version.

\subsection{Data Collection by Human Operators}

\leavevmode\hspace*{\parindent}%
We build controllable scenes in Unreal Engine and manually collect approximately 300 hours of human-operated interactive paired videos from 30+ Unreal scenes(Figure~\ref{fig:data_engine}). Each video has a resolution of 1920 $\times$ 1080 at 60 FPS. Rather than relying on scripted rollouts or autonomous agents, trained operators play every recorded trajectory, deliberately revisiting difficult viewpoints, interaction boundaries, transitions, and idle intervals. Operators are instructed to behave like real game players instead of executing isolated actions. Each trajectory contains complex combinations of movement and interaction inputs together with frequent changes in viewing direction.

Individual capture videos range from tens of minutes to hours in duration, providing extended trajectories that contain natural changes among exploration, interaction, camera movement, and idle behavior. The current data spans indoor and outdoor environments, varied spatial layouts, and basic gameplay states.

\subsection{Synchronized White-Box Rendering}

\leavevmode\hspace*{\parindent}%
Each scene has high-fidelity and white-box representations with matched spatial occupancy, collision boundaries, principal silhouettes, and interaction functions. The white-box representation removes final textures, materials, complex lighting, and nonessential geometric detail while preserving the structures and visible state changes that should constrain generation.

Both representations share the same timeline, character state, and camera configuration during capture. For every target frame, the data engine records the corresponding white-box frame at the same viewpoint and time. The paired streams carry different information. The white-box view specifies visible geometry, layout, occlusion, and principal motion. The high-fidelity target supplies appearance, illumination, material response, environmental detail, and secondary dynamics.

\subsection{Interaction Distribution}

\leavevmode\hspace*{\parindent}%
The collection includes interactions that expose visible consequences of engine-side gameplay, such as contact with walls and floors, traversal over stairs, changes in vehicle motion, blocking, sliding, and changes in character pose. They show how engine-resolved behavior should appear in the final visual stream.

Operators perform walking, running, jumping, climbing, camera rotation, driving, and sitting. The present collection prioritizes clean and verifiable coverage of locomotion, viewpoint change, vehicle control, simple state transitions, and idle behavior. Complex combat, multi-character cooperation, and dense object manipulation are not yet represented at comparable scale.

Idle behavior is equally important. Real play contains observation, hesitation, waiting, and intervals in which only the environment moves. A motion-heavy dataset can bias an autoregressive renderer toward unnecessary activity or unstable static scenes. We retain stationary characters, small camera adjustments, and environment-only motion so that the training distribution covers both action and the absence of action.

\subsection{Time-Aligned Structured Records}

\leavevmode\hspace*{\parindent}%
In addition to the two video streams, the data engine records camera pose, operator input, collisions, state transitions, and event information under a shared timestamp. These records preserve the provenance of each visual change and support data filtering and system analysis. The current version uses camera pose for visual-history retrieval, but operator input, collision, state-transition, and event fields do not participate in the training objective or condition interface.

\subsection{Appearance Annotation}

\leavevmode\hspace*{\parindent}%
Because physics, rules, and state transitions already exist in the engine, annotation is restricted to information absent from the white-box stream but required to establish visual identity. We use two fields: scene style and character appearance. Scene-style annotation summarizes the visual domain and art direction, while character-appearance annotation describes the visible identity and presentation of the controlled character. Neither field describes mechanics, physical rules, actions, or event outcomes.

We use Qwen3.6-27B to generate these annotations automatically from representative high-fidelity observations~\citep{qwen36_27b}. A fixed prompt produces the two fields separately and explicitly excludes gameplay semantics. The resulting text is aligned with the renderer's initialization role.

%% file: Sections/5_pipeline.tex
\section{Magpie System Architecture}
\label{sec:pipeline}

\begin{figure*}[t]
  \centering
  \includegraphics[width=0.97\textwidth]{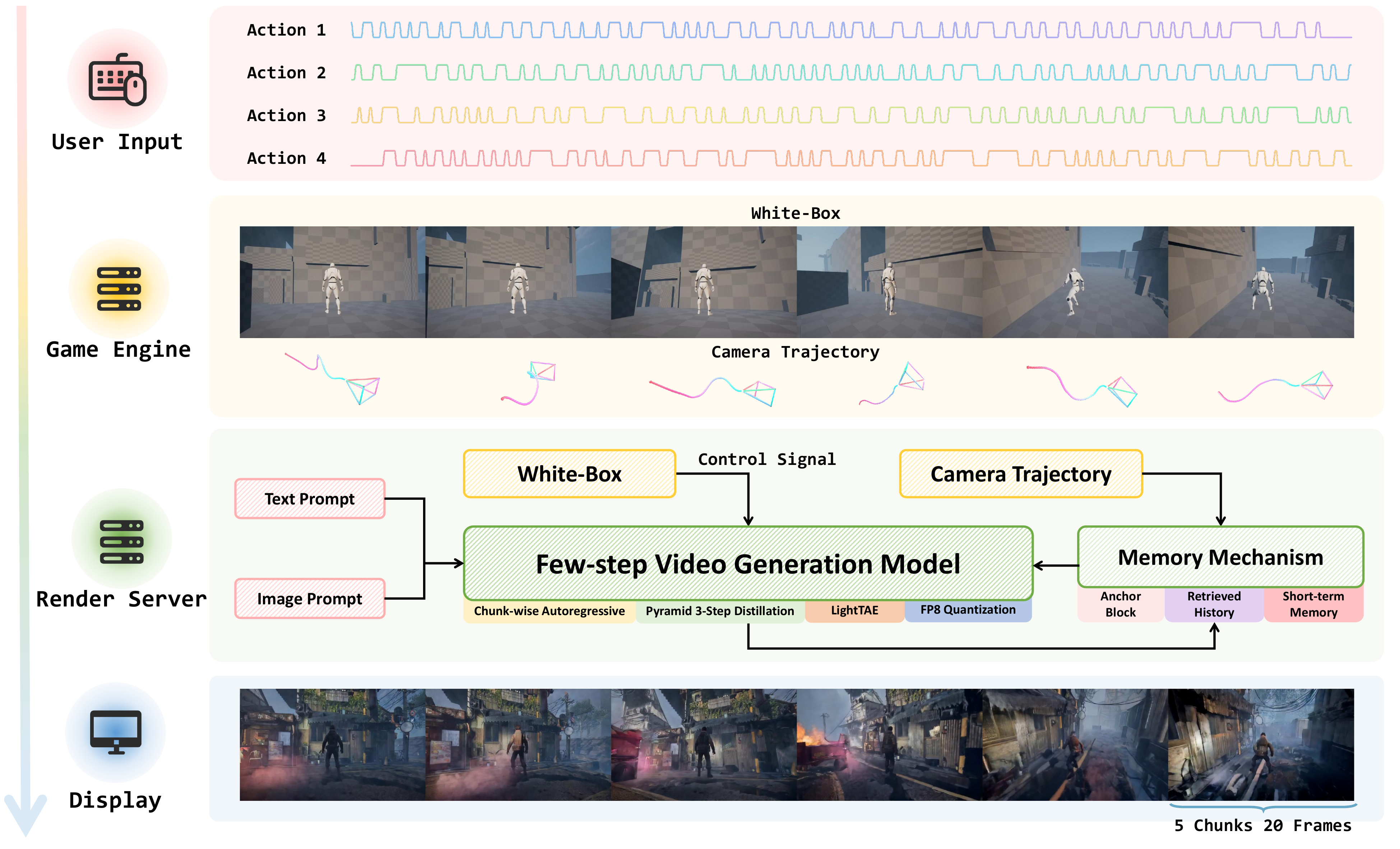}
  \caption{\textbf{Generative world-rendering pipeline.} User inputs are executed by the Game Engine, which produces synchronized white-box observations and camera trajectories. The Render Server initializes appearance from text and image prompts, uses white-box video as the continuing structural condition, and uses camera trajectory only to retrieve view-relevant history. The memory mechanism combines this retrieved history with an early anchor block and recent generated chunks. A chunk-wise few-step generator with pyramid distillation, LightTAE, and FP8 execution produces the displayed video.}
  \label{fig:renderer_pipeline}
\end{figure*}

\leavevmode\hspace*{\parindent}%
Magpie implements the architecture described in Section~\ref{sec:overview} as a distributed runtime composed of the Gameplay Design module, the User Client, the Game Engine, and an independent Render Server. Communication among the runtime components is implemented using WebRTC \citep{webrtc2025}. Conventional WebRTC jitter buffers are designed to maintain smooth playback under network variation, which can increase buffering delay. We instead configure the jitter buffer to prioritize frame freshness and low communication latency. Stale frames are discarded when necessary, allowing the effective frame rate to decrease rather than delaying subsequent frames. Magpie defines explicit message boundaries prevent gameplay state from bypassing the Game Engine or entering the generative model as an undeclared conditioning signal. This architecture modifies the visual path of a game prototype while retaining the Game Engine as the executable source of gameplay behavior.

\subsection{Gameplay Design Workflow}

\leavevmode\hspace*{\parindent}%
Gameplay Design constructs and maintains editable white-box scenes. Designers specify layouts, traversable regions, interactive rules and objects, physical parameters, triggers, and event logic, then establish the initial state of an interaction session.

Design metadata is not sent directly to the Render Server. An edit first changes the executable scene. The Game Engine then resolves the current state and produces an updated white-box observation. 

\subsection{User Client}

\leavevmode\hspace*{\parindent}%
The User Client displays generated video streaming and captures keyboard, mouse, or controller input. Each input is timestamped and forwarded to the Game Engine. The User Client does not modify world state locally and does not send controls to the Render Server. It only connects player input and generated output to the runtime loop.

\subsection{Game Engine}

\leavevmode\hspace*{\parindent}%
The Game Engine receives player input, applies the designed rules, and advances the gameplay state. It resolves traversal, collisions, interactions, events, and camera updates before producing the next white-box rendering. For each generation chunk, the Game Engine sends the Render Server two runtime products: the white-box observation and its camera pose. Object properties, actions, state variables, and event records remain inside the Game Engine. This boundary keeps stochastic visual synthesis separate from reproducible gameplay.

\subsection{Render Server}

\leavevmode\hspace*{\parindent}%
The Render Server manages the lifecycle of the generative world renderer. At session initialization, it accepts the text prompt and first-frame image. For every later chunk, it injects the engine-rendered white-box frame into denoising and uses the camera pose only to retrieve relevant history for visual consistency. It manages conditions and inference but does not make gameplay decisions.

\subsection{Closed-Loop Execution}

\leavevmode\hspace*{\parindent}%
Each interaction cycle follows one causal order. After the User Client captures an action, the Game Engine resolves its gameplay consequence and renders the resulting white-box observation with its camera pose. The Render Server then generates the visual segment presented by the User Client. The generative model therefore receives a visible result that has already been determined by gameplay execution, rather than predicting or redefining the rule-level outcome. The current implementation performs this exchange at chunk boundaries.

%% file: Sections/6_training.tex
\section{Generative World Renderer}
\label{sec:renderer}

\begin{figure*}[t]
  \centering
  \includegraphics[width=0.97\textwidth]{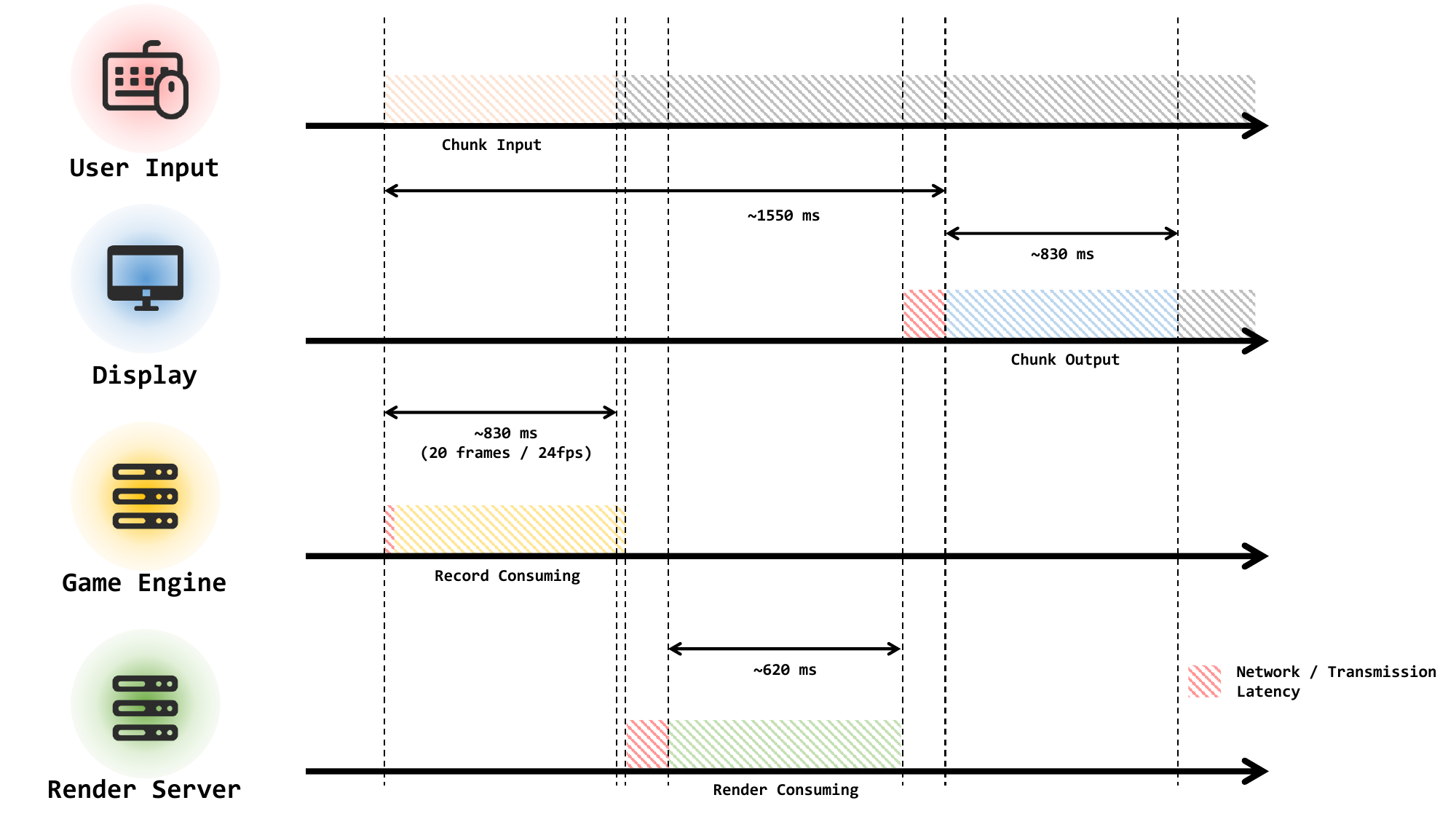}
  \caption{\textbf{Chunk-level interaction and rendering timeline.} At 24 FPS, the Game Engine records 20 white-box frames in approximately 830 ms. After transfer, the Render Server generates the corresponding chunk in approximately 620 ms; transfer and buffering account for the short intervals between stages. The first action-aligned visual response therefore appears after roughly 1550 ms, while subsequent output chunks are displayed over approximately 830 ms.}
  \label{fig:runtime_timeline}
\end{figure*}

\leavevmode\hspace*{\parindent}%
Our real-time renderer is built on Wan2.2-TI2V-5B backbone~\citep{wan2025}. Figure~\ref{fig:renderer_pipeline} summarizes its role in the end-to-end interaction path. To support long-form generation at interactive compute throughput, it adopts the bounded multi-scale context, hierarchical generation, and few-step distillation designs of Helios~\citep{helios2026}. Magpie introduces conditional control signals to the autoregressive video generator. Text and a first-frame style image initialize appearance, white-box frames provide continuing structural control, and camera poses are reserved for historical indexing. 

\subsection{Restricted Conditions}

\leavevmode\hspace*{\parindent}%
At initialization, the renderer receives a text prompt $c$ and first-frame image $r$. Together they establish scene style, character appearance, and the initial visual state. The initialization is encoded once as $s_0=E(c,r)$.

For each subsequent generation chunk (5 latents), the game engine produces a white-box observation $g_t$. This observation preserves visible scene layout, principal geometry, occlusion, character placement, and engine-resolved motion while omitting final materials, lighting, texture, and nonessential geometric detail. The renderer injects $g_t$ throughout denoising as its only continuing external condition.

We evaluate three mechanisms for injecting the white-box condition. These mechanisms should take care of both the controlled quality and the computational cost, since our ultimate goal is to distill the model for real‑time inference. The first embeds the encoded condition directly into the noisy latent. This design provides strong consistency to the white-box structure with minimal computational cost, but it also suppresses visual detail and causes a noticeable reduction in rendering quality. The second modulates intermediate features through adaptive layer normalization~\citep{peebles2023dit}. Although AdaLN conveys the condition effectively within an individual chunk, it produces pronounced visual discontinuities at chunk boundaries. The third places the encoded white-box representation in the history-conditioning sequence and introduces it into the denoising network through cross-attention. In our comparison, cross-attention provides the most practical balance, with acceptable overall quality and consistency to the white-box observation. We therefore adopt cross-attention as the condition injection mechanism in the current model.

Let $z_\tau$ denote the noisy latent for the current chunk, $h_t$ the assembled visual history, and $\tau$ the diffusion timestep. The denoising network is written as
\[
\epsilon_\theta = D_\theta(z_\tau,\tau \mid g_t,h_t,s_0).
\]
Player actions, game state, object properties, and event signals are currently absent from this function. They affect generation only indirectly. The engine-side variables may change the visible white-box observation, while camera pose affects which historical observations are placed in $h_t$. The restricted interface keeps the model focused on visual synthesis. 

\subsection{Autoregressive History Context}

\leavevmode\hspace*{\parindent}%
The renderer generates a long session as a sequence of short chunks. Each chunk is conditioned on the current white-box observation and a bounded visual history. The history is assembled in the following fixed order:

\paragraph{Early anchor block.} Chunks near the beginning of the session retain the appearance and scene identity established by the first-frame image. Keeping this block stable reduces long-horizon departure from the initialization.

\paragraph{Retrieved history.} Following Context-as-Memory~\citep{contextasmemory2025}, camera poses are used to compute field-of-view overlap and retrieve observations relevant to the current viewpoint. Position, orientation, and frustum participate only in this indexing procedure. They are not encoded as denoising conditions.

\paragraph{Recent generated chunks.} The latest outputs preserve local motion, short-term appearance, and continuity across adjacent chunks. Older recent chunks are evicted as new chunks are produced so that the active context remains bounded.

The resulting context is
\[
h_t=[h^{\mathrm{anchor}},h_t^{\mathrm{relevant}},h_t^{\mathrm{recent}}],
\]
the anchor stabilizes initialization, retrieved history supports revisitation, and recent history preserves local continuity.

\subsection{Hierarchical Few-Step Generation}

\leavevmode\hspace*{\parindent}%
We follow the hierarchical generation and multi-stage distillation procedure of Helios~\citep{helios2026}. Hierarchical generation first resolves coarse spatial structure and motion, then restores higher-frequency appearance. This allocation reduces the cost of applying every denoising operation at full detail.

Distillation progressively transfers a multi-step teacher trajectory to a 3-step denoising student. The student then reduces the discrepancy between training on clean video and inference on generated context by self forcing. Within Magpie, the objective is to preserve white-box condition adherence and the ordered history interface while using a few-step generator.

\subsection{Deployment Optimization}

\leavevmode\hspace*{\parindent}%
Runtime cost is further reduced through a lightweight autoencoder and mixed-precision execution. LightTAE~\citep{lightx2v2025} replaces a more expensive spatiotemporal autoencoder for latent encoding and decoding, operations that occur repeatedly at chunk boundaries and therefore contribute directly to interaction latency. The model backbone uses FP8 mixed precision where numerical sensitivity permits. Linear projections, attention operations, and feed-forward layers use low-precision computation, while normalization, temporal embeddings, and sensitive operators retain higher precision. 

Each Magpie inference step follows a fixed chunk-level schedule. The Render Server first encodes one chunk of white-box video transmitted by the Game Engine into latent conditions, then assembles the bounded visual history, denoises a single output chunk, and finally decodes the generated latents into video frames. To maintain stable computational cost and latency over arbitrarily long generation sessions, both encoding and decoding operate one chunk at a time, so their workloads do not grow with the duration of the session. 

%% file: Sections/7_performance.tex
\section{Performance Analysis}
\label{sec:performance}

\begin{figure*}[t]
  \centering
  \includegraphics[width=0.97\textwidth]{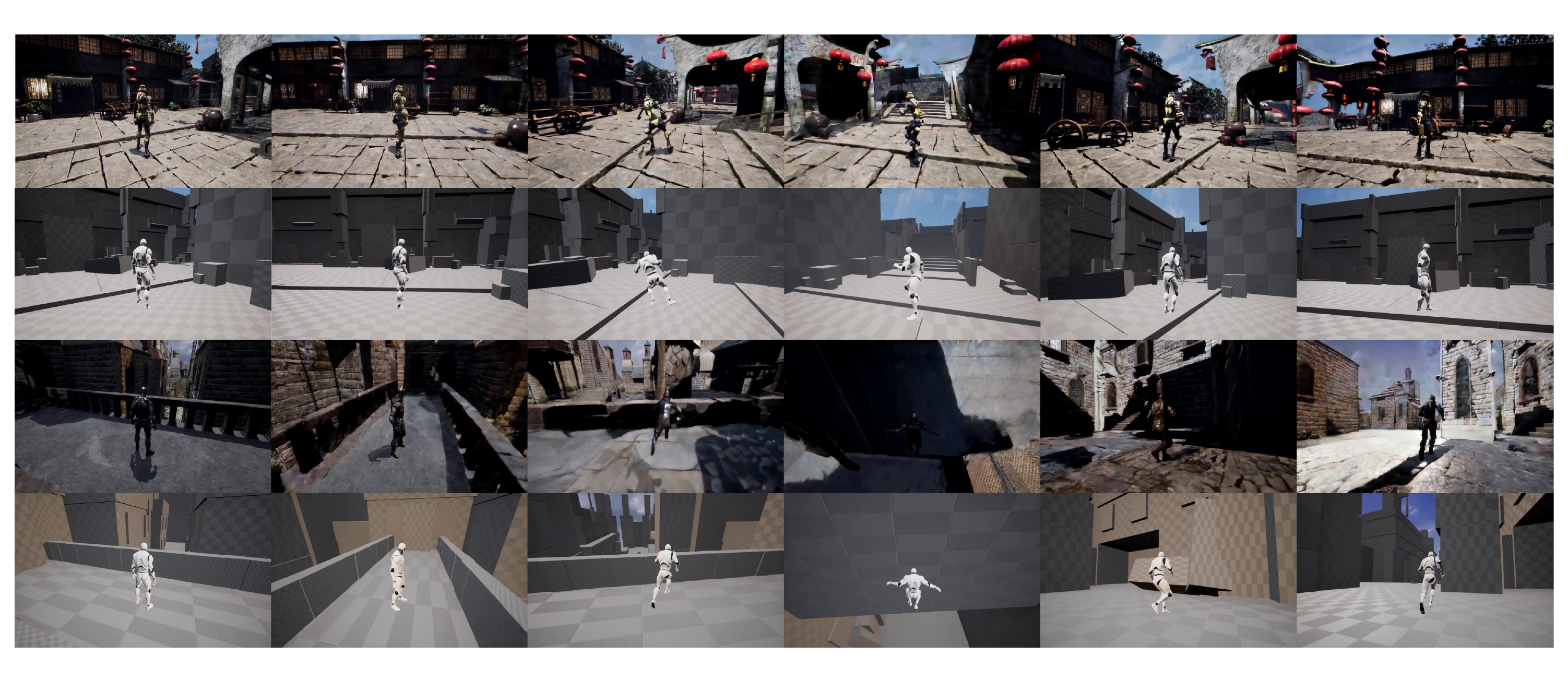}
  \caption{Qualitative results at $1280 \times 768$ produced by the base autoregressive model.}
  \label{fig:base_results}
\end{figure*}

\leavevmode\hspace*{\parindent}%
We analyze the runtime performance of the Magpie 5B Render Server on a single NVIDIA H100 GPU. Figure~\ref{fig:runtime_timeline} shows how engine capture, transmission, rendering, and display compose at chunk boundaries. We report compute-side throughput separately from end-to-end interaction latency because the former measures model execution after warm-up, whereas the latter also includes preparation of the engine condition. Inference proceeds autoregressively at the chunk level. Each chunk contains five latent frames. The first chunk is decoded into 17 video frames, while every subsequent chunk is decoded into 20 frames.

\subsection{Stable Chunk-Wise Decoding}

\leavevmode\hspace*{\parindent}%
The Render Server decodes only one chunk at a time to keep the decoding workload and inference latency stable throughout a long interaction session. Directly treating consecutive chunks as independent inputs is not compatible with the temporal causality of the Wan decoder because the decoding of the current chunk depends on preceding temporal context. Magpie therefore applies a bridge between adjacent chunks before decoding. The bridge supplies the causal context required at each chunk boundary while preserving a fixed chunk-wise decoding schedule. As a result, decoder latency does not grow with the length of the generated session.

\subsection{Render Server Throughput}

\leavevmode\hspace*{\parindent}%
After the runtime reaches a stable operating state, the total prediction time for one chunk is approximately 620 ms. Since a regular chunk produces 20 video frames, the generation rate on the compute side is approximately 32.2 FPS.
This rate measures the throughput of the Render Server after warm-up. It is higher than the 24 FPS rate used by the interaction pipeline, which allows the server to sustain the target video rate during stable chunk-wise generation. The first chunk uses a different decoding layout and produces 17 frames, so it is excluded from the steady-state throughput calculation.

The distilled 5B model reaches a peak GPU memory usage of approximately 34 GB during inference under this configuration. This measurement includes the memory required by the deployed inference pipeline at runtime and characterizes the single-GPU resource requirement alongside its steady-state throughput.

\begin{table}[t]
  \centering
  \small
  \begin{tabular}{lc}
    \toprule
    Configuration & Value \\
    \midrule
    Model size & 5B \\
    GPU & 1$\times$ NVIDIA H100 \\
    Inference mode & Chunk-wise autoregressive \\
    Decoding strategy & Chunk-wise with bridging \\
    Resolution & $1280 \times 768$ \\
    Latents per chunk & 5 \\
    First-chunk output & 17 frames \\
    Subsequent-chunk output & 20 frames \\
    Stable prediction time & 620 ms per chunk \\
    Compute-side throughput & 32.2 FPS \\
    Peak GPU memory usage & 34 GB \\
    \bottomrule
  \end{tabular}
  \caption{Steady-state inference performance of the Magpie 5B Render Server.}
  \label{tab:render_server_performance}
\end{table}

\subsection{End-to-End Interaction Latency}

\leavevmode\hspace*{\parindent}%
Compute throughput does not directly determine interaction latency because the Render Server cannot begin a chunk until the corresponding white-box video sequence is available. The delay from a user action to the corresponding generated observation consists of three stages.

\paragraph{White-box sequence preparation.} The Game Engine first executes the user action and prepares the white-box video sequence required by one inference chunk. A regular chunk corresponds to 20 frames. At 24 FPS, collecting this sequence takes approximately 0.83s.
\paragraph{Transmission and encoding.} The white-box sequence and camera metadata are transferred to the Render Server and encoded for inference. This stage contributes approximately 0.1 s and is small relative to chunk preparation and generation.

\paragraph{Generative rendering.} The Render Server predicts and decodes the output chunk in approximately 620 ms under stable operation.

The end-to-end response latency can be summarized as
\[
T_{\mathrm{response}}
=T_{\mathrm{engine}}+T_{\mathrm{transmission}}+T_{\mathrm{render}}
\approx 1.55\ \mathrm{s}.
\]
Magpie therefore begins to provide visual feedback corresponding to a newly received user action after approximately 1.55 seconds.

\begin{table}[t]
  \centering
  \small
  \begin{tabular}{lc}
    \toprule
    Latency component & Duration \\
    \midrule
    Game Engine white-box chunk preparation & 0.83 s \\
    Network transmission and encoding & $\sim$0.1 s \\
    Render Server inference and decoding & 0.62 s \\
    \midrule
    End-to-end response & $\sim$1.55 s \\
    \bottomrule
  \end{tabular}
  \caption{Latency from user input to the corresponding generated visual feedback.}
  \label{tab:end_to_end_latency}
\end{table}

%% file: Sections/8_results.tex
\subsection{Qualitative Results}
\label{sec:results}

\leavevmode\hspace*{\parindent}%

\begin{figure*}[t]
  \centering
  \includegraphics[width=0.97\textwidth]{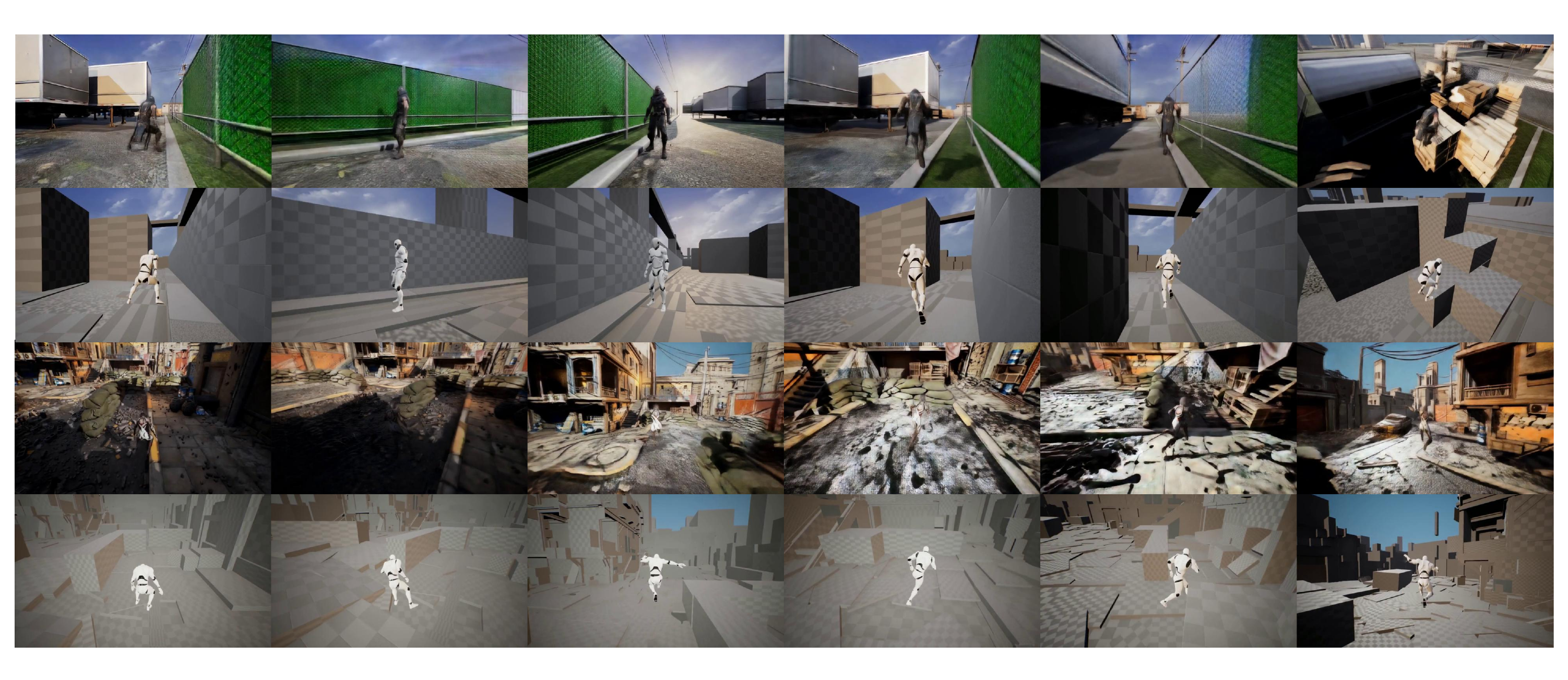}
  \caption{Qualitative results at $1280 \times 768$ produced by the distilled real-time renderer.}
  \label{fig:distill_results}
\end{figure*}

\leavevmode\hspace*{\parindent}%
We present qualitative results of the base renderer and the distilled real-time renderer in Figures~\ref{fig:base_results} and~\ref{fig:distill_results}. We think the evaluation of generative renderer should focus on three aspects of the generated video, including long-horizon memory, consistency with the white-box condition, and overall visual quality.

%% file: Sections/9_conclusion.tex
\section{Limitations}
\label{sec:limitations}

\leavevmode\hspace*{\parindent}%
The current system demonstrates that engine-owned gameplay can be combined with generative visual rendering, but it does not yet meet the latency, fidelity, and persistence requirements of a production game renderer. 

\paragraph{Latency from white-box pre-recording and chunk-wise inference.}
The Game Engine currently records a complete white-box chunk before the Render Server begins the corresponding generation step. At 24 FPS, preparing 20 white-box frames alone takes approximately 830 ms, after which chunk-wise transfer, inference, and decoding introduce additional delay. This latency of magpie is approximately 1.6 s, which is too high for interactions that require immediate feedback. A frame-wise streaming architecture is therefore necessary. White-box frames should be transferred and consumed as soon as they are produced, allowing engine execution, condition encoding, generation, and display to overlap instead of waiting at chunk boundaries.

\paragraph{Depth ambiguity in white-box conditions.}
White-box rendering preserves visible layout, silhouettes, occlusion, and motion, but a single RGB observation can remain ambiguous about metric depth, surface orientation, thin structures, and the spatial relationship between similarly colored regions. These ambiguities can cause the generated image to violate scene geometry even when its two-dimensional structure appears plausible. A richer condition interface could supplement the white-box stream with engine-derived depth, normals, segmentation, motion vectors, or other geometric buffers. Depth is a particularly direct extension because it can disambiguate foreground--background relationships without transferring gameplay rules or hidden state into the renderer.

\paragraph{Inconsistency to white-box conditions.}
Even when the white-box observation contains the required visible structure, the renderer does not follow it with uniform reliability. Under rapid motion, large viewpoint changes, occlusion, complex interactions, or scenes outside the strongest part of the training distribution, generated geometry, character placement, and object boundaries may drift away from the engine observation. Improving consistency will require stronger condition encoding and injection, training objectives that explicitly penalize structural deviation, broader coverage of difficult transitions, and runtime consistency checks capable of detecting or correcting outputs that conflict with the white-box input.

\paragraph{Insufficient high-fidelity appearance supervision.}
The current paired engine captures provide accurate synchronization and controllable structural supervision, but their visual targets do not fully cover the diversity and quality of real-world materials, lighting, weather, human motion, and fine-scale physical effects. This limitation constrains the photorealism and robustness of the learned renderer. Future training should combine the current paired data with both high-quality real-world video and high-quality rendered results, including offline or production-grade renderings with richer assets, lighting, and effects. Real video can provide natural appearance and motion priors, while high-quality rendered data can offer cleaner control over scene content and may retain geometric metadata that supports alignment with white-box conditions. Because these sources do not always include directly aligned white-box observations, the mixture may require additional rendering passes, geometric condition construction, pseudo-labeling, or staged pretraining and fine-tuning. The balance must be controlled carefully so that richer supervision improves visual quality without weakening adherence to the Game Engine's visible structure.

\paragraph{Absence of audio generation and synchronization.}
Magpie currently produces only the visual observation. In a complete interactive experience, sound effects, ambience, dialogue, and music must remain synchronized with engine-resolved events and generated imagery. The Game Engine can provide authoritative audio events or conventional audio playback, but the visual Render Server may still require additional optimization to maintain audiovisual alignment, especially when generated motion changes the apparent timing or character of a visible event. Supporting audio therefore requires both a defined engine-to-audio interface and tighter control of rendering latency and temporal synchronization.

\paragraph{Lack of persistent three-dimensional visual memory.}
The current memory mechanism combines an early anchor block, FOV-retrieved observations, and recent generated chunks. This bounded two-dimensional history supports short-term continuity and revisitation, but it does not maintain an explicit, persistent representation of the generated world. Consequently, appearance may drift across large viewpoint changes or long absences, and previously rendered content must be inferred again from incomplete observations. A stronger system should continuously project generated observations into an offline three-dimensional representation that can be updated, stored, and reused during later rendering. Such a representation could provide view-consistent appearance and geometry while keeping the Game Engine as the authority for gameplay state.

\paragraph{High inference cost and limited edge deployability.}
The current distilled 5B renderer remains computationally expensive, requiring a high-end server GPU and reaching approximately 34 GB of peak GPU memory during inference. This resource requirement prevents practical deployment on most client and edge devices and increases the infrastructure cost of serving interactive sessions. Future work should develop substantially smaller and more efficient models through improved architectures, distillation, quantization, and deployment-aware optimization, with the goal of moving generative rendering from a dedicated Render Server to local edge hardware while preserving structural adherence, temporal stability, and acceptable visual quality.

\paragraph{Frame rate and visual quality.}
The reported 32.2 FPS still unable to meet the requirements of the games. Moreover, few-step generation, low-precision execution, bounded context, and limited training coverage introduce a visible quality gap relative to the intended renderer. Current outputs can exhibit temporal instability, texture degradation, identity drift, and inconsistency to engine-visible detail. 

\section{Conclusion}
\label{sec:conclusion}

\leavevmode\hspace*{\parindent}%
Magpie provides a system-level implementation path for introducing generative video into real-time game rendering without replacing the engine-side structures that make gameplay designable and reproducible. Designers construct scenes and rules in the game engine, the Game Engine resolves actions and maintains world state, and the independent Render Server converts the resulting white-box observations into generated video. The model is therefore responsible for visual presentation, while the engine remains responsible for gameplay execution and state control. This separation allows a playable white-box scene to support richer materials, lighting, environmental detail, motion, and visual styles before complete production assets are available. For players, it creates the possibility of more flexible and personalized presentation. For developers, it reduces the dependence of early prototypes on a complete visual-asset pipeline and allows gameplay and experience design to be evaluated earlier. More broadly, the system suggests that generative models in games need not replace the engine. Instead, they could serve as the bridge.

%% file: main.bbl
\begin{thebibliography}{51}
\providecommand{\natexlab}[1]{#1}

\bibitem[{Agarwal et~al.(2026)Agarwal, Ali, Allen, Antolini, Aubame, Azzolini,
  Bai, Bala, Balaji, Bapst et~al.}]{agarwal2026cosmos}
Agarwal, N.; Ali, A.; Allen, J.; Antolini, M.; Aubame, A.; Azzolini, A.; Bai,
  J.; Bala, M.; Balaji, Y.; Bapst, J.; et~al. 2026.
\newblock Cosmos 3: Omnimodal world models for physical ai.
\newblock \emph{arXiv preprint arXiv:2606.02800}.

\bibitem[{Bian et~al.(2026)Bian, Xue, Zhang, Zhang, Jin, Li, Zhuang, Li, Huang,
  Huang, Duan, and Xu}]{echoinfinity2026}
Bian, Y.; Xue, Z.; Zhang, S.; Zhang, S.; Jin, W.; Li, Y.; Zhuang, J.; Li, H.;
  Huang, J.; Huang, H.; Duan, N.; and Xu, Q. 2026.
\newblock Echo-Infinity: Learning Evolving Memory for Real-Time Infinite Video
  Generation.
\newblock \emph{arXiv preprint arXiv:2606.04527}.

\bibitem[{Chen et~al.(2026{\natexlab{a}})Chen, Lin, Yang, Zhang, Fei, Li, Chen,
  Ao, Pang, Wang et~al.}]{chen2026skyreels}
Chen, G.; Lin, D.; Yang, J.; Zhang, Y.; Fei, Z.; Li, D.; Chen, S.; Ao, C.;
  Pang, N.; Wang, Y.; et~al. 2026{\natexlab{a}}.
\newblock Skyreels-v4: Multi-modal video-audio generation, inpainting and
  editing model.
\newblock \emph{arXiv preprint arXiv:2602.21818}.

\bibitem[{Chen et~al.(2026{\natexlab{b}})Chen, Li, Kong, Zhu, Xu, Xiao, Guo,
  Ye, Zhang, Zhao et~al.}]{chen2026unividx}
Chen, H.; Li, H.; Kong, X.; Zhu, T.; Xu, S.; Xiao, W.; Guo, Y.; Ye, C.; Zhang,
  L.; Zhao, H.; et~al. 2026{\natexlab{b}}.
\newblock UniVidX: A Unified Multimodal Framework for Versatile Video
  Generation via Diffusion Priors.
\newblock \emph{arXiv preprint arXiv:2605.00658}.

\bibitem[{Chen et~al.(2026{\natexlab{c}})Chen, Wang, Huang, Yang, Zhang, Xiao,
  Chu, Mao, Hu, Liu et~al.}]{chen2026longlive2}
Chen, Y.; Wang, L.; Huang, W.; Yang, S.; Zhang, B.; Xiao, Y.; Chu, R.; Mao, W.;
  Hu, Q.; Liu, S.; et~al. 2026{\natexlab{c}}.
\newblock LongLive-2.0: An NVFP4 Parallel Infrastructure for Long Video
  Generation.
\newblock \emph{arXiv preprint arXiv:2605.18739}.

\bibitem[{{DreamX Team} et~al.(2026){DreamX Team}, Bai, Chen, Chu, Dang, Dou,
  Gao, Gu, Hong, Lei et~al.}]{dreamxworld2026}
{DreamX Team}; Bai, Y.; Chen, R.; Chu, X.; Dang, R.; Dou, H.; Gao, B.; Gu, Q.;
  Hong, S.; Lei, J.; et~al. 2026.
\newblock DreamX-World 1.0: A General-Purpose Interactive World Model.
\newblock \emph{arXiv preprint arXiv:2606.16993}.

\bibitem[{Gao et~al.(2026)Gao, Wang, Zhu, Chen, Liu, Bai, Wang, Yuan, Wang, Lu,
  Cheng, Zhang, Gao, Feng, Liu, Yao, Xu, Zhu, Shen, and
  Ouyang}]{lingbotworld2_2026}
Gao, Z.; Wang, Q.; Zhu, J.; Chen, J.; Liu, Z.; Bai, Q.; Wang, J.; Yuan, Y.;
  Wang, H.; Lu, Y.; Cheng, K.~L.; Zhang, H.; Gao, J.; Feng, T.; Liu, Y.; Yao,
  Y.; Xu, Y.; Zhu, X.; Shen, Y.; and Ouyang, H. 2026.
\newblock Infinite Worlds with Versatile Interactions.
\newblock \emph{arXiv preprint arXiv:2607.07534}.

\bibitem[{Geng et~al.(2025)Geng, Deng, Bai, Kolter, and He}]{geng2026mean}
Geng, Z.; Deng, M.; Bai, X.; Kolter, Z.; and He, K. 2025.
\newblock Mean flows for one-step generative modeling.
\newblock \emph{Advances in Neural Information Processing Systems}, 38:
  75460--75482.

\bibitem[{Gomez-Nogales et~al.(2026)Gomez-Nogales, Hong, Ge, Zhuang,
  Comino-Trinidad, Casas, and Zhou}]{coarsetoreal2026}
Gomez-Nogales, G.; Hong, Y.; Ge, C.; Zhuang, P.; Comino-Trinidad, M.; Casas,
  D.; and Zhou, Y. 2026.
\newblock Coarse-to-Real: Generative Rendering for Populated Dynamic Scenes.
\newblock \emph{arXiv preprint arXiv:2601.22301}.

\bibitem[{Gu et~al.(2026)Gu, Fang, Jiang, Mao, Han, Cai, and
  Shou}]{anyflow2026}
Gu, Y.; Fang, G.; Jiang, Y.; Mao, W.; Han, S.; Cai, H.; and Shou, M.~Z. 2026.
\newblock AnyFlow: Any-Step Video Diffusion Model with On-Policy Flow Map
  Distillation.
\newblock \emph{arXiv preprint arXiv:2605.13724}.

\bibitem[{Gu et~al.(2025)Gu, Yan, Lu, Li, Dou, Si, Dong, Liu, Lin, Liu, Wang,
  and Liu}]{diffusionasshader2025}
Gu, Z.; Yan, R.; Lu, J.; Li, P.; Dou, Z.; Si, C.; Dong, Z.; Liu, Q.; Lin, C.;
  Liu, Z.; Wang, W.; and Liu, Y. 2025.
\newblock Diffusion as Shader: 3D-Aware Video Diffusion for Versatile Video
  Generation Control.
\newblock \emph{arXiv preprint arXiv:2501.03847}.

\bibitem[{Guo et~al.(2026)Guo, Litman, He, Miller, Chen, and
  Ramanan}]{guo2026lightcrafter}
Guo, Z.; Litman, Y.; He, Y.; Miller, J.; Chen, C.; and Ramanan, D. 2026.
\newblock LightCrafter: PBR-Conditioned Video Diffusion Refinement for
  Controllable and Consistent Relighting.
\newblock \emph{arXiv preprint arXiv:2607.08016}.

\bibitem[{HaCohen et~al.(2026)HaCohen, Brazowski, Chiprut, Bitterman, Kvochko,
  Berkowitz, Shalem, Lifschitz, Moshe, Porat et~al.}]{hacohen2026ltx}
HaCohen, Y.; Brazowski, B.; Chiprut, N.; Bitterman, Y.; Kvochko, A.; Berkowitz,
  A.; Shalem, D.; Lifschitz, D.; Moshe, D.; Porat, E.; et~al. 2026.
\newblock Ltx-2: Efficient joint audio-visual foundation model.
\newblock \emph{arXiv preprint arXiv:2601.03233}.

\bibitem[{He et~al.(2025)He, Peng, Liu, Wang, Zhang, Cui, Kang, Jiang, An, Ren
  et~al.}]{he2025matrix2}
He, X.; Peng, C.; Liu, Z.; Wang, B.; Zhang, Y.; Cui, Q.; Kang, F.; Jiang, B.;
  An, M.; Ren, Y.; et~al. 2025.
\newblock Matrix-game 2.0: An open-source real-time and streaming interactive
  world model.
\newblock \emph{arXiv preprint arXiv:2508.13009}.

\bibitem[{Hong et~al.(2025)Hong, Mei, Ge, Xu, Zhou, Bi, Hold-Geoffroy, Roberts,
  Fisher, Shechtman et~al.}]{hong2025relic}
Hong, Y.; Mei, Y.; Ge, C.; Xu, Y.; Zhou, Y.; Bi, S.; Hold-Geoffroy, Y.;
  Roberts, M.; Fisher, M.; Shechtman, E.; et~al. 2025.
\newblock Relic: Interactive video world model with long-horizon memory.
\newblock \emph{arXiv preprint arXiv:2512.04040}.

\bibitem[{Hu et~al.(2026{\natexlab{a}})Hu, Volhejn, Rahary, Mulder, Makkar,
  Liao, Royer, Orsini, Jelley, Alonso et~al.}]{hu2026mira}
Hu, A.; Volhejn, V.; Rahary, A.~R.; Mulder, C.; Makkar, A.; Liao, A.; Royer,
  A.; Orsini, M.; Jelley, A.; Alonso, E.; et~al. 2026{\natexlab{a}}.
\newblock Multiplayer Interactive World Models with Representation
  Autoencoders.
\newblock \emph{arXiv preprint arXiv:2607.05352}.

\bibitem[{Hu et~al.(2026{\natexlab{b}})Hu, Wang, Zhang, and
  Zhang}]{hu2026wantofight}
Hu, L.; Wang, G.; Zhang, P.; and Zhang, B. 2026{\natexlab{b}}.
\newblock WanToFight: Real-Time Generative Game Engine for Multi-Player Combat
  Interaction.
\newblock \emph{arXiv preprint arXiv:2607.12592}.

\bibitem[{Huang et~al.(2025)Huang, Li, He, Zhou, and
  Shechtman}]{selfforcing2025}
Huang, X.; Li, Z.; He, G.; Zhou, M.; and Shechtman, E. 2025.
\newblock Self Forcing: Bridging the Train-Test Gap in Autoregressive Video
  Diffusion.
\newblock \emph{arXiv preprint arXiv:2506.08009}.

\bibitem[{Huang et~al.(2026)Huang, Wang, Tan, Yu, Zhang, Zheng, Liu, Chuang,
  and Zhang}]{generativeworldrenderer2026}
Huang, Z.-H.; Wang, Z.; Tan, J.; Yu, R.; Zhang, Y.; Zheng, B.; Liu, Y.-L.;
  Chuang, Y.-Y.; and Zhang, K. 2026.
\newblock Generative World Renderer.
\newblock \emph{arXiv preprint arXiv:2604.02329}.

\bibitem[{Jennings et~al.(2025)Jennings, Castelli, Bostr{\"o}m, and
  Bruaroey}]{webrtc2025}
Jennings, C.; Castelli, F.; Bostr{\"o}m, H.; and Bruaroey, J.-I. 2025.
\newblock {WebRTC}: Real-Time Communication in Browsers.
\newblock W3c recommendation, World Wide Web Consortium.

\bibitem[{Jiang et~al.(2026)Jiang, Sun, Wang, Zhu, Wang, Zhang, Liu, Wang,
  Zheng, Sun et~al.}]{jiang2026abot}
Jiang, F.; Sun, Z.; Wang, M.; Zhu, Z.; Wang, C.; Zhang, Y.; Liu, W.; Wang, Y.;
  Zheng, X.; Sun, R.; et~al. 2026.
\newblock ABot-World-0: Infinite Interactive World Rollout on a Single Desktop
  GPU.
\newblock \emph{arXiv preprint arXiv:2607.19191}.

\bibitem[{Liang et~al.(2025)Liang, Gojcic, Ling, Munkberg, Hasselgren, Lin,
  Gao, Keller, Vijaykumar, Fidler et~al.}]{liang2025diffusionrenderer}
Liang, R.; Gojcic, Z.; Ling, H.; Munkberg, J.; Hasselgren, J.; Lin, C.-H.; Gao,
  J.; Keller, A.; Vijaykumar, N.; Fidler, S.; et~al. 2025.
\newblock Diffusionrenderer: Neural inverse and forward rendering with video
  diffusion models.
\newblock In \emph{2025 IEEE/CVF Conference on Computer Vision and Pattern
  Recognition (CVPR)}, 26069--26080. IEEE.

\bibitem[{{LightX2V Contributors}(2025)}]{lightx2v2025}
{LightX2V Contributors}. 2025.
\newblock {LightX2V}: Light Video Generation Inference Framework.
\newblock \url{https://github.com/ModelTC/lightx2v}.

\bibitem[{Moxin et~al.(2026)Moxin, Ji, Chen, Zhang, Yang, Zhu, Zhao, Xie, Wang,
  Liu et~al.}]{moxin2026moworld}
Moxin, T.; Ji, D.; Chen, T.; Zhang, X.; Yang, J.; Zhu, Q.; Zhao, A.; Xie, Z.;
  Wang, H.; Liu, X.; et~al. 2026.
\newblock MoWorld: A Flash World Model.
\newblock \emph{arXiv preprint arXiv:2607.06216}.

\bibitem[{Nam et~al.(2026)Nam, Hong, Huang, Liu, Lee, Kim, Jin, Lee, Jung,
  Choi, Kim, and Zhou}]{worldcam2026}
Nam, J.; Hong, Y.; Huang, C.-H.~P.; Liu, F.; Lee, J.; Kim, J.; Jin, S.; Lee,
  Y.; Jung, J.; Choi, S.; Kim, S.; and Zhou, Y. 2026.
\newblock WorldCam: Interactive Autoregressive 3D Gaming Worlds with Camera
  Pose as a Unifying Geometric Representation.
\newblock \emph{arXiv preprint arXiv:2603.16871}.

\bibitem[{Peebles and Xie(2023)}]{peebles2023dit}
Peebles, W.; and Xie, S. 2023.
\newblock Scalable Diffusion Models with Transformers.
\newblock In \emph{Proceedings of the IEEE/CVF International Conference on
  Computer Vision (ICCV)}, 4195--4205.

\bibitem[{{Qwen Team}(2026)}]{qwen36_27b}
{Qwen Team}. 2026.
\newblock {Qwen3.6-27B}: Flagship-Level Coding in a 27B Dense Model.

\bibitem[{Sauer et~al.(2024)Sauer, Lorenz, Blattmann, and
  Rombach}]{sauer2024adversarial}
Sauer, A.; Lorenz, D.; Blattmann, A.; and Rombach, R. 2024.
\newblock Adversarial diffusion distillation.
\newblock In \emph{European Conference on Computer Vision}, 87--103. Springer.

\bibitem[{Savva et~al.(2026)Savva, Michel, Lu, Waiwitlikhit, Meehan, Mishra,
  Poddar, Lu, and Xie}]{savva2026solaris}
Savva, G.; Michel, O.; Lu, D.; Waiwitlikhit, S.; Meehan, T.; Mishra, D.;
  Poddar, S.; Lu, J.; and Xie, S. 2026.
\newblock Solaris: Building a multiplayer video world model in minecraft.
\newblock \emph{arXiv preprint arXiv:2602.22208}.

\bibitem[{Sun et~al.(2025)Sun, Zhang, Wang, Wu, Wang, Wang, Wang, Zhang, Wang,
  and Guo}]{worldplay2025}
Sun, W.; Zhang, H.; Wang, H.; Wu, J.; Wang, Z.; Wang, Z.; Wang, Y.; Zhang, J.;
  Wang, T.; and Guo, C. 2025.
\newblock WorldPlay: Towards Long-Term Geometric Consistency for Real-Time
  Interactive World Modeling.
\newblock \emph{arXiv preprint arXiv:2512.14614}.

\bibitem[{Team et~al.(2026)Team, Gao, Wang, Zeng, Zhu, Cheng, Li, Wang, Xu, Ma
  et~al.}]{team2026lingbot1}
Team, R.; Gao, Z.; Wang, Q.; Zeng, Y.; Zhu, J.; Cheng, K.~L.; Li, Y.; Wang, H.;
  Xu, Y.; Ma, S.; et~al. 2026.
\newblock Advancing open-source world models.
\newblock \emph{arXiv preprint arXiv:2601.20540}.

\bibitem[{{Team Seedance} et~al.(2026)}]{seedance2026}
{Team Seedance}; et~al. 2026.
\newblock Seedance 2.0: Advancing Video Generation for World Complexity.
\newblock \emph{arXiv preprint arXiv:2604.14148}.

\bibitem[{{Team Wan} et~al.(2025){Team Wan}, Wang, Ai, Wen et~al.}]{wan2025}
{Team Wan}; Wang, A.; Ai, B.; Wen, B.; et~al. 2025.
\newblock Wan: Open and Advanced Large-Scale Video Generative Models.
\newblock \emph{arXiv preprint arXiv:2503.20314}.

\bibitem[{Wang et~al.(2026{\natexlab{a}})Wang, Zhang, Kabra, Uijlings,
  Waslander, Zisserman, Carreira, He, Andriluka, Bazavan
  et~al.}]{wang2026video}
Wang, L.; Zhang, C.; Kabra, R.; Uijlings, J.; Waslander, S.; Zisserman, A.;
  Carreira, J.; He, K.; Andriluka, M.; Bazavan, E.~G.; et~al.
  2026{\natexlab{a}}.
\newblock Video Generation Models are General-Purpose Vision Learners.
\newblock \emph{arXiv preprint arXiv:2607.09024}.

\bibitem[{Wang et~al.(2025)Wang, Luo, Shi, Jia, Lu, Xue, Wang, Wan, Zhang, and
  Gai}]{cinemaster2025}
Wang, Q.; Luo, Y.; Shi, X.; Jia, X.; Lu, H.; Xue, T.; Wang, X.; Wan, P.; Zhang,
  D.; and Gai, K. 2025.
\newblock CineMaster: A 3D-Aware and Controllable Framework for Cinematic
  Text-to-Video Generation.
\newblock \emph{arXiv preprint arXiv:2502.08639}.

\bibitem[{Wang et~al.(2026{\natexlab{b}})Wang, Zhao, Yang, Chen, Zhang, He,
  Duan, Chen, Yang, and Zhuang}]{latentspatialmemory2026}
Wang, W.; Zhao, H.; Yang, Y.; Chen, F.; Zhang, Z.; He, Y.; Duan, Z.; Chen,
  D.~Y.; Yang, Y.; and Zhuang, B. 2026{\natexlab{b}}.
\newblock Latent Spatial Memory for Video World Models.
\newblock \emph{arXiv preprint arXiv:2606.09828}.

\bibitem[{Wang et~al.(2026{\natexlab{c}})Wang, Liu, Li, Huang, Xu, Kang, An,
  Wang, Jiang, Wei et~al.}]{matrixgame3_2026}
Wang, Z.; Liu, Z.; Li, J.; Huang, K.; Xu, B.; Kang, F.; An, M.; Wang, P.;
  Jiang, B.; Wei, Y.; et~al. 2026{\natexlab{c}}.
\newblock Matrix-Game 3.0: Real-Time and Streaming Interactive World Model with
  Long-Horizon Memory.
\newblock \emph{arXiv preprint arXiv:2604.08995}.

\bibitem[{Xin et~al.(2026)Xin, Priyadarshi, Xin, Kartal, Vavre, Thekkumpate,
  Chen, Mahabaleshwarkar, Shahaf, Bercovich et~al.}]{xin2026quantization}
Xin, M.; Priyadarshi, S.; Xin, J.; Kartal, B.; Vavre, A.; Thekkumpate, A.~K.;
  Chen, Z.; Mahabaleshwarkar, A.~S.; Shahaf, I.; Bercovich, A.; et~al. 2026.
\newblock Quantization-aware distillation for nvfp4 inference accuracy
  recovery.
\newblock \emph{arXiv preprint arXiv:2601.20088}.

\bibitem[{Xiong et~al.(2026)Xiong, Song, Kang, Yan, Jiang, Yang, Fu, Fotiadis,
  Wang, Liu et~al.}]{xiong2026actworld}
Xiong, Z.; Song, Y.; Kang, H.; Yan, Q.; Jiang, L.; Yang, J.; Fu, Z.; Fotiadis,
  S.; Wang, A.; Liu, Z.; et~al. 2026.
\newblock ActWorld: From Explorable to Interactive World Model via Action-Aware
  Memory.
\newblock \emph{arXiv preprint arXiv:2606.17730}.

\bibitem[{Xu et~al.(2026)Xu, Jiang, Shu, Sunkavalli, Patel, and
  Mei}]{wonder2026}
Xu, J.; Jiang, H.; Shu, Z.; Sunkavalli, K.; Patel, V.~M.; and Mei, Y. 2026.
\newblock Wonder: Video World Model Done Better.
\newblock \emph{arXiv preprint arXiv:2607.26037}.

\bibitem[{Yang et~al.(2025)Yang, Huang, Chu, Xiao, Zhao, Wang, Li, Xie, Chen,
  Lu, Han, and Chen}]{longlive2025}
Yang, S.; Huang, W.; Chu, R.; Xiao, Y.; Zhao, Y.; Wang, X.; Li, M.; Xie, E.;
  Chen, Y.; Lu, Y.; Han, S.; and Chen, Y. 2025.
\newblock LongLive: Real-Time Interactive Long Video Generation.
\newblock \emph{arXiv preprint arXiv:2509.22622}.

\bibitem[{Yi et~al.(2026)Yi, Kim, Cho, Jang, Yun, and Kim}]{worldkv2026}
Yi, J.; Kim, M.; Cho, P.~H.; Jang, W.; Yun, S.; and Kim, S. 2026.
\newblock WorldKV: Efficient World Memory with World Retrieval and Compression.
\newblock \emph{arXiv preprint arXiv:2605.22718}.

\bibitem[{Yin et~al.(2024{\natexlab{a}})Yin, Gharbi, Park, Zhang, Shechtman,
  Durand, and Freeman}]{yin2024improved}
Yin, T.; Gharbi, M.; Park, T.; Zhang, R.; Shechtman, E.; Durand, F.; and
  Freeman, W.~T. 2024{\natexlab{a}}.
\newblock Improved distribution matching distillation for fast image synthesis.
\newblock \emph{Advances in neural information processing systems}, 37:
  47455--47487.

\bibitem[{Yin et~al.(2024{\natexlab{b}})Yin, Gharbi, Zhang, Shechtman, Durand,
  Freeman, and Park}]{yin2024dmd}
Yin, T.; Gharbi, M.; Zhang, R.; Shechtman, E.; Durand, F.; Freeman, W.~T.; and
  Park, T. 2024{\natexlab{b}}.
\newblock One-step diffusion with distribution matching distillation.
\newblock In \emph{2024 IEEE/CVF Conference on Computer Vision and Pattern
  Recognition (CVPR)}, 6613--6623. IEEE.

\bibitem[{Yu et~al.(2025)Yu, Bai, Qin, Liu, Wang, Wan, Zhang, and
  Liu}]{contextasmemory2025}
Yu, J.; Bai, J.; Qin, Y.; Liu, Q.; Wang, X.; Wan, P.; Zhang, D.; and Liu, X.
  2025.
\newblock Context as Memory: Scene-Consistent Interactive Long Video Generation
  with Memory Retrieval.
\newblock \emph{arXiv preprint arXiv:2506.03141}.

\bibitem[{Yuan et~al.(2026)Yuan, Yin, Li, Huang, Yang, and Yuan}]{helios2026}
Yuan, S.; Yin, Y.; Li, Z.; Huang, X.; Yang, X.; and Yuan, L. 2026.
\newblock Helios: Real Real-Time Long Video Generation Model.
\newblock \emph{arXiv preprint arXiv:2603.04379}.

\bibitem[{Zhang et~al.(2025)Zhang, Peng, Wang, Wang, Zhu, Kang, Jiang, Gao, Li,
  Liu, and Zhou}]{matrixgame2025}
Zhang, Y.; Peng, C.; Wang, B.; Wang, P.; Zhu, Q.; Kang, F.; Jiang, B.; Gao, Z.;
  Li, E.; Liu, Y.; and Zhou, Y. 2025.
\newblock Matrix-Game: Interactive World Foundation Model.
\newblock \emph{arXiv preprint arXiv:2506.18701}.

\bibitem[{Zhao et~al.(2026)Zhao, Zhu, Zheng, Zhou, Yan, Li, Yang, Li, and
  Zhu}]{causalforcingplusplus2026}
Zhao, M.; Zhu, H.; Zheng, K.; Zhou, Z.; Yan, B.; Li, X.; Yang, X.; Li, C.; and
  Zhu, J. 2026.
\newblock Causal Forcing++: Scalable Few-Step Autoregressive Diffusion
  Distillation for Real-Time Interactive Video Generation.
\newblock \emph{arXiv preprint arXiv:2605.15141}.

\bibitem[{Zheng et~al.(2026)Zheng, Wang, Ma, Chen, Zhang, Balaji, Chen, Liu,
  Zhu, and Zhang}]{zheng2026rcm}
Zheng, K.; Wang, Y.; Ma, Q.; Chen, H.; Zhang, J.; Balaji, Y.; Chen, J.; Liu,
  M.-Y.; Zhu, J.; and Zhang, Q. 2026.
\newblock Large scale diffusion distillation via score-regularized
  continuous-time consistency.
\newblock In \emph{International Conference on Learning Representations
  (ICLR)}.

\bibitem[{Zhu et~al.(2026{\natexlab{a}})Zhu, Liu, Zhao, Ye, Chen, Yu, He, Han,
  and Xie}]{zhu2026sana}
Zhu, H.; Liu, H.; Zhao, Y.; Ye, T.; Chen, J.; Yu, J.; He, T.; Han, S.; and Xie,
  E. 2026{\natexlab{a}}.
\newblock Sana-wm: Efficient minute-scale world modeling with hybrid linear
  diffusion transformer.
\newblock \emph{arXiv preprint arXiv:2605.15178}.

\bibitem[{Zhu et~al.(2026{\natexlab{b}})Zhu, Zhao, He, Su, Li, and
  Zhu}]{causalforcing2026}
Zhu, H.; Zhao, M.; He, G.; Su, H.; Li, C.; and Zhu, J. 2026{\natexlab{b}}.
\newblock Causal Forcing: Autoregressive Diffusion Distillation Done Right for
  High-Quality Real-Time Interactive Video Generation.
\newblock \emph{arXiv preprint arXiv:2602.02214}.

\end{thebibliography}
